\documentclass[submission,copyright,creativecommons]{eptcs}
\usepackage{amsmath}
\usepackage{amssymb}
\usepackage{amsfonts}
\usepackage{mathtools}
\usepackage{mathptmx}
\usepackage{listings}
\usepackage{tikz}
\usepackage{graphicx}
\usepackage{subfig}
\usepackage{tikzscale}
\usepackage{float}
\usepackage{backnaur}
\usepackage{alltt}

\newfloat{listing}{htbp}{lop}
\floatname{listing}{Listing}
\newenvironment{bnfsplit}[1][0.7\textwidth]
{\minipage[t]{#1}$}
{$\endminipage}

\renewenvironment{alltt}{\vspace{-0.75\baselineskip}\begin{oldalltt}}{\end{oldalltt}\vspace{-0.1\baselineskip}}

\graphicspath{{images/}{tikz/}}

\title{Generating Local Search Neighborhood \\ with Synthesized Logic Programs}
\author{Mateusz \'Sla\.zy\'nski \qquad 
    \institute{AGH University of Science\\ and Technology, Poland}
    \email{mslaz@agh.edu.pl} 
    \and
    Salvador Abreu
    \institute{University of \'Evora and LISP, \\ Portugal}
    \email{spa@uevora.pt}
    \and Grzegorz J. Nalepa
    \institute{AGH University of Science\\ and Technology, Poland}
    \email{gjn@agh.edu.pl}
}
\def\titlerunning{Generating Local Search Neighborhood with Synthesized Logic 
Programs}
\def\authorrunning{M. \'Sla\.zy\'nski, S. Abreu and G. J. Nalepa}

\newif\ifcomment\commentfalse

\begin{document}

\label{firstpage}

\maketitle

  \begin{abstract}
    Local Search meta-heuristics have been proven a viable approach to
    solve difficult optimization problems.  Their performance depends
    strongly on the search space landscape, as defined by a cost
    function and the selected neighborhood operators.  In this paper
    we present a logic programming based framework, named Noodle,
    designed to generate bespoke Local Search neighborhoods tailored
    to specific discrete optimization problems.  The proposed system
    consists of a domain specific language, which is inspired by logic
    programming, as well as a genetic programming solver, based on the
    grammar evolution algorithm.  We complement the description with a
    preliminary experimental evaluation, where we synthesize efficient
    neighborhood operators for the traveling salesman problem, some of
    which reproduce well-known results.
  \end{abstract}


\section{Introduction}
%

Size and high dimensionality of the combinatorial problems often make them
infeasible for the exhaustive optimization methods, thus giving rise
to meta-heuristics --- a family of methods aimed at efficiently
exploring the search space in pursuit of good enough local optima.
Local Search algorithms --- well established members of this family
--- represent the search space in terms of the optimization criterion
and a so-called neighborhood operator, which defines a neighborhood
relation over the candidate solutions (configurations).  A Local
Search solver starts with a single configuration and then replaces it
with one of its neighbors, repeating this process until the
termination condition is met.  The high impact of the neighborhood
operator choice on the solvers' performance has induced vast amounts
of research on problem-specific operators, which often outperform the
generic search
strategies~\cite{van_hentenryck_traveling_2006,januario_new_2016}.

While there has been noticeable recent progress in automated algorithm
configuration, including such complicated tasks as finding efficient
hyper-parameters~\cite{shahriari_taking_2016}, feature
selection~\cite{kaul_autolearn_2017} and designing deep neural network
architectures~\cite{elsken_neural_2019}, there is still no satisfying
method to create a problem-specific Local Search strategy.  This paper
aims at filling this gap by presenting a system capable of inferring
Local Search neighborhoods from a fine-grained problem representation
consisting in an augmented Constraint Programming model.

Section~\ref{sec:ndl} shortly presents a Neighborhood Description
Language~\cite{slazynski_towards_2019} --- a formalism designed to
express neighborhood operators in a declarative manner along with a
runtime environment.  The rest of the paper is dedicated to
synthesizing the aforementioned operators by means of a Grammar
Evolution~\cite{oneil_grammatical_2003} algorithm applied on a Logic Programming 
DSL tailored to the
given optimization problem.  The synthesis process is guided by a
fitness function which considers both syntax of the generated
program and various features of the actual induced neighborhood.  We
conclude the paper with initial experimental results, conducted on the
Traveling Salesman Problem, showing that our system is able to
``reinvent'' state-of-the-art neighborhood operators.


\section{Related Works}

Automatic inference of the search algorithm from a CP model was
already proposed in~\cite{elsayed_synthesis_2011}.  Authors of
Comet~\cite{van_hentenryck_constraint-based_2005} have proposed method
of synthesizing Local Search
strategies~\cite{van_hentenryck_synthesis_2007}.  The Comet modeling
language included mechanisms to independently model both a problems'
structure and the search strategy, including simple neighborhoods.
One of the inferred search strategy features was to use neighborhood
operators to replace some hard constraints, e.g.~the
\textit{alldifferent} constraint.  This approach was further
proclaimed in~\cite{he_solution_2012} and implemented by the
Yuck~\cite{marte_yuck_2017} and
OscaR/CBLS~\cite{bjordal_constraint-based_2015} solvers, providing
dedicated neighborhoods to handle specific global constraints. 
Recently, a system was proposed which infers plausible neighborhood
operators from data structures occurring in the Constraint Programming
model, as defined in the Essence language
~\cite{akgun_framework_2018}.  During the search, the solver switches
between those promising neighborhoods using classic multi-arm bandit
algorithms.  The resulting strategy proved to be superior to or
competitive with popular generic strategies on several common
combinatorial problems.


\section{The NDL Language}
\label{sec:ndl}

All experiments described herein, and their respective results, are
founded on a novel method of representing Local Search neighborhood
operators called the \emph{Neighborhood Definition Language} (NDL).
NDL uses the underlying CSP (or COP)\footnote{Constraint Satisfaction
  Problem or Constraint Optimisation Problem.} structure, stated as a
Constraint Programming model, to specify a non-deterministic program,
capable of inducing a set of new configurations (neighbors) starting
from a single point in the search space.

To capture the high-level structure of the considered problem, we have
extended the traditional constraint graph representation with labels,
which group constraints and variables based on the context of their
occurrence in the model.  This structure is called the \emph{Typed
  Constraint Graph} (or TCG), as the labels are called respectively
variable and constraint types:
\begin{itemize}

\item variable types
  $T_V = \lbrace t_{1}^{v}, t_{2}^{v}, \dots t_{n}^{v}\rbrace$
  correspond to the indexed data structures, i.e. arrays, commonly
  used in Constraint modeling languages.  Every constraint graph node
  $v \in V$ is then labeled after its origin array and the index it
  was defined with: $l_v: V \rightarrow \mathbb{I} \times T_V$, where
  $\mathbb{I}$ is a set of array indexes.

\item constraint types
  $T_C = \lbrace t_{1}^{c}, t_{2}^{c}, \dots t_{m}^{c}\rbrace$ capture
  the relation between constraints which share not only their
  semantics, but also their origin.  We say that constraints share
  their type if they have been defined by means of a \label{key}single
  aggregation function (e.g. quantifier) or they come from
  decomposition of a single global constraint.  Every constraint graph
  edge $c \in C$ is labeled with its respective type:
  $l_c: C \rightarrow T_C$.  It is worth mentioning that a \emph{Typed
    Constraint Graph} requires global constraints to be decomposed as
  it supports only binary constraints.

\end{itemize} 

Such a rich problem representation allows NDL programs to query the
configuration based both on its semantic and syntactic structure by
means of so called \emph{selectors}.  A selector is a
non-deterministic operator which can find a set of variables
satisfying specified constraints, e.g.~variables of a given type,
variables constrained by a specific type of constraint.  The
\emph{selected} variables may be then tested and accepted by
\emph{filters} --- basic arithmetic and logical tests allowing to
prune out unwanted variables, based on their values.  Finally
\textit{modifiers} are used to perturb a configuration by reassigning
new values to a given set of variables.  All of these operations may
be chained by means of functional composition and result in a
first-order non-deterministic program meant to produce sets of
neighboring configurations.

First-order NDL programs are limited in terms of expressiveness due to
fixed number of variables involved in every operator.  To amend this
lack, we augment the language with second-order \textit{combinators}.
As in recursion schemes found in functional programming
languages~\cite{meijer_functional_1991}, combinators perform common
primitively recursive computational patterns, lifting the first-order
programs to operate sequentially on several different parts of the
configuration.  There are three combinators available in the language:
\begin{itemize}
\item \textit{Selector Quantifier} applies a program to all possible
  results of a non-deterministic selector.
\item \textit{The Least Fixpoint Operator} iterates trough the Typed
  Constraint Graph edges in a breadth-first search manner and applies
  a given program to the edges nodes.
\item \textit{Iterator} treats a variable as the beginning of an ordered collection 
where every element is defined by the previous element's value; then
  performs the specified program on each of the collection's elements,
  similarly to \textit{map} in functional programming languages.
\end{itemize}

Due to the problems being of finite size and intrinsically limited
recursive nature, all combinators are bound to eventually finish
computation, thus eschewing Turing completeness and associated issues.
A more detailed list of NDL features and its components can be found
on the project wiki\footnote{See: https://gitlab.geist.re/pro/ndl/wikis/home} 
and 
in the language grammar
defined in Section~\ref{sub:grammar}.

\subsection{Noodle Runtime}
\label{sub:runtime}

\begin{figure}
  \centering
  \includegraphics[width=0.7\linewidth]{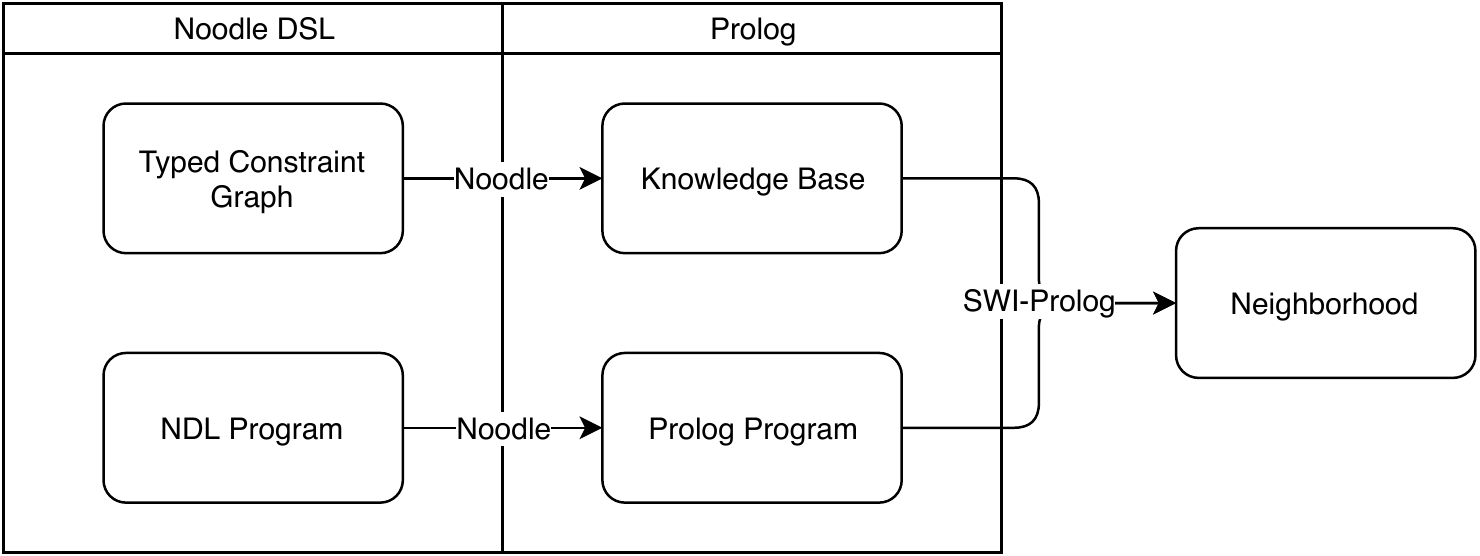}
  \caption{Evaluation of NDL program by the Noodle runtime.}
  \label{fig:noodle}
\end{figure}

At present, we have one implementation of the NDL language named
Noodle, built with the SWI-Prolog~\cite{wielemaker_swi-prolog_2012}
system.  The core of the Noodle runtime consists of two domain
specific languages, Noodle Domain Language (NoodleDL) and Noodle Query
Language (NoodleQL) respectively aimed at representing Typed
Constraint Graph and NDL programs.  Both languages are implemented by
means of Prolog meta-programming facilities and are compiled ahead of
time to plain Prolog: NoodleDL creates a knowledge base which encodes
the problem and NoodleQL is first type checked and then translated to
a program which implements the neighborhood operator per se.
Figure~\ref{fig:noodle} presents this process in a graphical manner.

The choice of Logic Programming for the NDL implementation was
motivated by several unique features.  Foremost, NDL depends on
non-deterministic evaluation, inherent to Logic Programming in the
form of backtracking.  Secondly, having the Typed Constraint Graph
encoded in the knowledge base, selectors may be efficiently
implemented as standard queries, leaving aside various technical
issues.  Last, but not the least, meta-programming features allowed us
to quickly prototype and improve the Noodle DSLs.

\section{Synthesis Algorithm}
\label{sec:algo}

A common method of synthesizing logic programs entails resorting to an
Inductive Logic Programming system, which creates programs given a
prior background knowledge and learning examples.  In our situation,
this approach cannot be applied as there is no learning data available
before finishing the synthesis process; in other words, induction
would be a viable method to find theories (operators) to define
a-priori known neighborhoods, but is unable to generate novel
operators.  Other possible approaches consist in the application of
discrete optimization methods, most notably evolutionary techniques
branded as Genetic Programming algorithms.  Unfortunately most of
those methods tend to be limited in terms of the program
representation, often allowing only explicit tree structures as used
in the LISP-like languages.  Another limitation, critical due to the
typed nature of NDL, is the so called type closure --- making all
values in the synthesized program share a single type.

To address these issues, the NDL framework uses the Grammar
Evolution~\cite{oneil_grammatical_2003} algorithm, a genetic programming 
method based
solely on the formal grammar of the language under consideration
paired with an adequate fitness function.  We now present a mechanism
for translating an NDL problem into a corresponding formal grammar,
exploiting structural information present in the model.  Next, we will
describe various components of the fitness function, together with an
experimental evaluation.

\subsection{Grammar}
\label{sub:grammar}

As previously stated, we chose to build the NDL framework on a Prolog
runtime engine.  Knowing this, the grammar which we'll use to
introduce the formal NDL language will be naturally expressible as a
Logic program.  Every neighborhood operator will thus be represented
as a conjunction of atoms:

\begin{bnf*}
    \bnfprod{program}
    {\bnfpn{atom} \bnfor \bnfpn{atom} \bnfsp \bnftd{$\land$} \bnfsp 
    \bnfpn{program}}
\end{bnf*}

There are four types of atoms in every program: 
\begin{bnf*}
    \bnfprod{atom}
    {\bnfpn{selector} \bnfor \bnfpn{modifier} \bnfor \bnfpn{filter} \bnfor 
    \bnfpn{combinator}}
\end{bnf*}

Selectors are used to query the NDL model for specific variables,
constraints and values occurring in the CSP being represented.  The
number and shape of the selectors depends therefore on the types
existing in the model.  Selectors are the sources of the operator:
they bring in parts of the CSP, designating sets of constraints,
variables or values.

\begin{bnf*}
  \bnfprod{selector} %
  { \begin{bnfsplit} %
      %
      constraint(c, \bnfpn{t1}, \bnfpn{t2})%
      %
      \bnfor variable(\bnfpn{i}, \bnfpn{t}) \\ %
      %
      \bnfor value(\bnfpn{t}, \bnfpn{d}) %
      %
      \bnfor constant(c, \bnfpn{d})
    \end{bnfsplit}}
\end{bnf*}

Variables in the problem may have their value modified via one of the
three available modifiers: \textit{set} --- setting the specified
value, \textit{swap} --- swapping two variables' values and
\textit{flip} --- flipping variable's value between two possible
outcomes.  Considering variables of type t and domain d:

\begin{bnf*}
  \bnfprod{modifier} { \begin{bnfsplit} %
      %
      set(\bnfpn{t}, \bnfpn{d}) \bnfor swap(\bnfpn{t}, \bnfpn{t})
      \bnfor flip(\bnfpn{t}, \bnfpn{d}, \bnfpn{d})
    \end{bnfsplit}}
\end{bnf*}

Filters are used to prune out unwanted results by checking basic
properties of the current configuration, e.g.~whether a given
constraint is satisfied or whether two values are equal.

\begin{bnf*}
  \bnfprod{filter}%
  { \begin{bnfsplit} %
      %
      is\_satisfied(c, \bnfpn{t1}, \bnfpn{t2}) %
      \bnfor is\_violated(c, \bnfpn{t1}, \bnfpn{t2}) \\ %
      %
      \bnfor \bnfpn{t}\bnfsp\bnftd{$\neq$}\bnfpn{t}
      \bnfor \bnfpn{t}\bnfsp\bnftd{$=$}\bnfsp\bnfpn{t} %
      %
      \bnfor \bnfpn{d}\bnfsp\bnftd{$\neq$}\bnfpn{d} \bnfor
      \bnfpn{d}\bnfsp\bnftd{$=$}\bnfsp\bnfpn{d} \bnfor
      \bnfpn{d}\bnfsp\bnftd{$<$}\bnfsp\bnfpn{d}
    \end{bnfsplit}}
\end{bnf*}

Our grammar includes three types of second-order atoms \textit{for\_each}, 
\textit{bfs\_over} and \textit{iterate} corresponding adequately to \textit{Selector 
Quantifier}, \textit{The Least Fixpoint Operator} and \textit{Iterator} introduced in 
Section~\ref{sec:ndl}.
The only addition are two variations: \textit{bfs\_over\_inverted} exploring the 
constraint graph with inverted arcs and \textit{iterate\_reversed} that treats a 
given variable as the last element of the considered collection.
It is worth mentioning that \textit{iterate} (and \textit{iterate\_reversed}) may be 
only used when the variable's 
domain and index
set coincide:

\begin{bnf*}
  \bnfprod{combinator}%
  { \begin{bnfsplit} %
      for\_each(\bnfpn{selector}, \bnfpn{program}) \\
      %
      \bnfor bfs\_over(c, \bnfpn{t1}, \bnfpn{t1}-\bnfpn{t2}, \bnfpn{program}) \\
      \bnfor bfs\_over\_inverted(c, \bnfpn{t2}, \bnfpn{t1}-\bnfpn{t2},
      \bnfpn{program}) \\
      %
      \bnfor iterate(\bnfpn{t}, \bnfpn{t}-\bnfpn{t}), \bnfpn{program}) \\
      \bnfor iterate\_reversed(\bnfpn{t}, \bnfpn{t}-\bnfpn{t}), \bnfpn{program}) \\
    \end{bnfsplit}}
\end{bnf*}

Finally, in order to synthesize atoms' arguments we have to introduce
terminals for each variable type in the NDL model.
For each NDL type, there are three sets of attributes we're interested
in, those which represent NDL variables (\textit{t}), indexes
(\textit{i}) and values (\textit{d}).

\begin{bnf*}
  \bnfprod{t}{ Tt0 \bnfor Tt1 \bnfor \dots \bnfor TtN}\\        
  \bnfprod{d}{Dd0 \bnfor Dd1 \bnfor \dots \bnfor DdM}\\
  \bnfprod{i}{Ii0 \bnfor Ii1 \bnfor \dots \bnfor IiK}
\end{bnf*}

It is worth noting that when a variable's domain and index set are
congruent, the corresponding \textit{i} and \textit{d} attributes may
also collapse to a single non-terminal.

The basic grammar already covers the core concepts, but it may be
augmented with a few features, aiming to make it more expressive:
\begin{itemize}

\item Local Scopes --- NDL language features local variables bound
  only in the local scope of a combinator.
  We partition the sets of non-terminals corresponding to the
  variables into two sets of global and local, constraining the second
  one to occur only inside the local scopes.

\item Symmetry Breaking --- the basic grammar is highly susceptible to
  a very simple symmetry --- variables' names are only labels and may
  be permuted, like colors in the graph coloring problem. While BNF grammars can 
  not impose any order on the used non-terminals, some of the variables still may 
  be fixed in certain situations, e.g. in the first atom of the program.
  
\item Limited Depth --- the presented grammar allows one to synthesize nested
  combinators, leading to arbitrarily large and thus slow programs.
  In order to prevent this behavior, one can prohibit such nested combinators at 
  the grammar level.

\end{itemize}

\subsection{Fitness Function}
\label{sub:fitness}

Given the formal grammar, we are able to synthesize syntactically
correct neighborhood operators.  Every generated program is then
evaluated by the Noodle runtime, resulting in several metrics which
may later be used by the genetic algorithm's fitness function.  The
evaluation is done in two phases:

\paragraph{Static Code Analysis} Before executing a program, Noodle
performs static analysis on the code, identifying semantically
incorrect or unused atoms.  Such atoms are treated as ineffective (so
called `introns') and are removed from the code only for time of execution.  
This has beneficial effect both on the execution phase, which does not have to 
consider runtime errors, and for the genetic algorithm itself as introns are
believed to improve the algorithm's
efficiency~\cite{angeline_advances_1994}. 

This phase results in four metrics:

\begin{itemize}
\item \textbf{used outputs ratio}
  $\textit{R}_o \in \left(0, 1 \right)$, i.e.~the number
  of atoms (e.g. selectors) whose outputs are used by another term.
\item \textbf{provided inputs ratio}
  $\textit{R}_i\in \left(0, 1\right)$, i.e.~the number
  of correctly used atoms that require input variables to be bound
  (e.g. filters and modifiers).
\item \textbf{unique arguments ratio}
  $\textit{R}_u\in \left( 0, 1 \right)$ counting atoms with no repeating arguments.
\item \textbf{effective atoms ratio}
  $\textit{R}_e\in \left( 0, 1 \right)$ is the most
  ``global'' metric --- it relates to number of atoms that have any
  effect on the configuration, either directly (by being a properly
  stated modifier) or indirectly (by providing input to a modifier).
\end{itemize}

All code related metrics pertain to the number of possible errors
and, in the best case scenario, will have value $1$ to indicate the
inexistence of any issues.  We decided to use a basic fitness
function, equal to arithmetic average of the above metrics:

$$\phi_\textit{code} = \left(\textit{R}_o + \textit{R}_i + \textit{R}_u + 
  \textit{R}_e\right)/4$$

It is worth noting that while this $\phi_\textit{code}$ does not say
anything about the neighborhood itself, it guides the search process
into areas containing a higher density of `healthy' individuals
(correct programs.)  This has a positive impact, especially on the
early generations, which otherwise contain mostly random and incorrect
programs.

\paragraph{Neighborhood Analysis} After pruning out the introns, the
neighborhood operator is compiled to an executable program\footnote{At
  this time, it's a regular Prolog program.} which is applied on a
testing configuration, in order to sample its neighborhood
$N = \left\lbrace n_0, n_1, \dots, n_k \right\rbrace$.
For every neighbor $n \in N$ we calculate two types of partial
metrics:

\begin{itemize}
\item \textbf{changes ratio}
  $\textit{R}_\textit{ch}: N \rightarrow \left( 0, 1\right)$, which
  accounts for the number of differences (variables with different
  values) between $n$ and the initial configuration.
\item \textbf{satisfied constraints ratio}
  $\textit{R}_\textit{sat}: N \times T_C \rightarrow \left( 0,
    1\right)$, that counts the satisfied constraints for a specific
  constraint type, in neighbor $n$.  It is computed for every
  constraint type in the model.
\end{itemize}

Those values are then accumulated into four neighborhood metrics:

\begin{itemize}
\item Size of the neighborhood $s = \vert N \vert$.
\item Number of unique neighbors $u \leq s$.
\item Statistics on introduced changes: $ch_{min}$, $ch_{max}$,
  $ch_{avg}$, $ch_{stdev}$, respectively the minimum, maximum, average
  and standard deviation of set
  $\lbrace R_{ch}(n) \mid n \in N \rbrace$.
\item Statistics on satisfied constraints: $sat_{min}^t$,
  $sat_{max}^t$, $sat_{avg}^t$, $sat_{stdev}^t$, respectively the
  minimum, maximum, average and standard deviation of set
  $\lbrace R_{sat}(n,t) \mid n \in N \rbrace$, for every constraint
  type $t \in T_C$.
\end{itemize}

As there are many ways to evaluate a neighborhood, we propose several
candidate measures based on the metrics we just presented.  The
following functions were designed to promote neighborhoods that keep
Local Search in a search sub-space which preserves the admissibility
of configurations, i.e.~one that keeps all the constraints satisfied.
Another motivation is to find diversified neighborhoods, with a
reasonable proportion of unique configurations:

\begin{itemize}

\item \textbf{size} score
  $\phi_\textit{size} = \frac{1}{1 + e^{\alpha_s \times (-u +
      \beta_s/2)}}$, which promotes larger neighborhoods, but flattens
  after a specified point due to the logistic function; in our
  experimentation, parameters $\alpha_s = 0.5$ resulted in a good
  slope and
  $\beta_s = \frac{\vert V \vert!}{2 \times (\vert V \vert -2)!}$ has
  been used to promote neighborhood operators which manipulate at
  least two different variables.

\item ratio of \textbf{unique neighbors vs.~total neighborhood size}
  $\phi_\textit{unique} = \frac{u}{s}$.  This function penalizes
  neighborhoods which contain many duplicates.

\item \textbf{normalized mean squared satisfiability} score
  $\phi_\textit{NMSS} = \frac{\sum\limits_{t \in T_C}
    (sat_{min}^t)^2}{\vert T_C \vert \times \vert V \vert \times
    chg_{avg}}$, which rewards neighborhoods in proportion to the
  number of satisfied constraints for each constraint type.  It is
  normalized against the number of changes in the neighborhood, to
  prevent bias toward operators which don't introduce any changes to
  the configuration.

\item \textbf{satisfiability} score
  $\phi_\textit{sat} = \frac{\sum\limits_{t \in T_C} score(t)}{\vert
    T_C \vert}$, which rewards the neighborhoods that satisfy as many
  constraints as possible.  The following $score$ function is designed
  to prefer neighborhoods that contain even single promising
  neighbors:
  \[   
    \textit{score}(t) = 
    \begin{cases}
      sat_{min}^t = 1, & 1 \\
      {otherwise}, & \frac{sat_{max}^t}{\vert V \vert}
    \end{cases}
  \] 

\item \textbf{variability} score
  $\phi_\textit{var} = \frac{1}{1 + e^{\alpha_v \times (-ch_{stdev} +
      \beta_v)}}$, which promotes neighborhoods in which
  configurations differ in a varying number of variables.  It may be
  applied to look after `complex' neighborhood operators.  In our
  experiments, we have set parameters $\alpha_v = 40$ and
  $\beta_v = 0.06$ in order to flatten this function quickly.
\end{itemize}


\section{Experimental Results}
\label{sec:experiment}

In order to evaluate the Noodle framework, we have conducted several
experiments of synthesizing neighborhood operators for the Traveling
Salesman Problem.  This problem has been chosen because of its
popularity and vast body of research on efficient local search
strategies~\cite{johnson_traveling_1997}.

\subsection{Experimental Setup}

\begin{figure}
    \centering
    \includegraphics[width=1\linewidth]{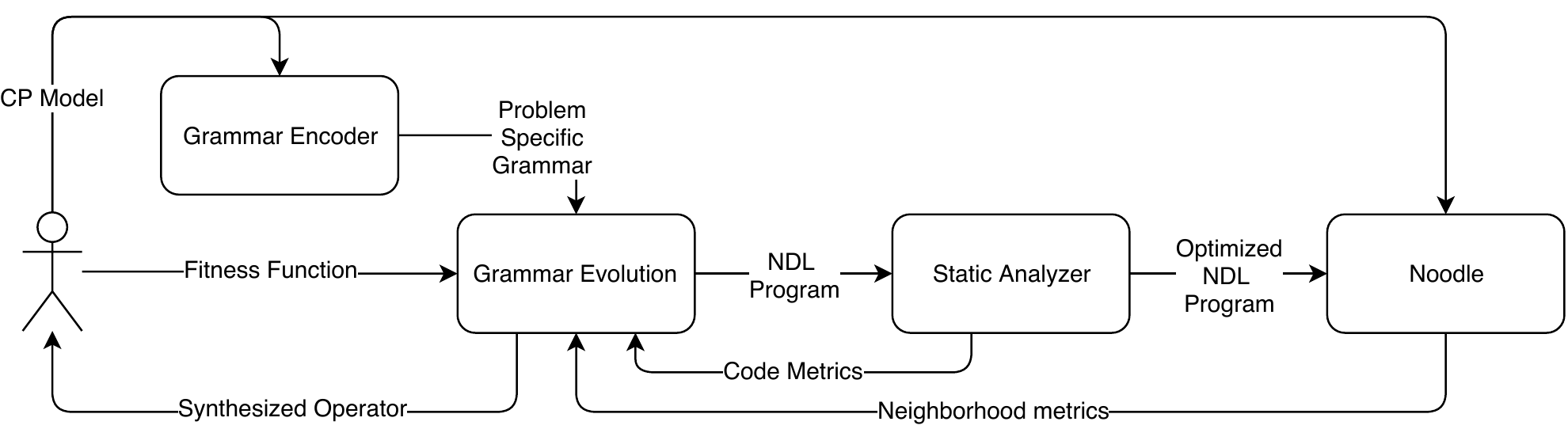}
    \caption{Experiment architecture.}
    \label{fig:experiment}
\end{figure}

Grammar Evolution can be tuned using various hyper-parameters specific
to the genetic programming algorithms.  The results we present here
were achieved using default settings, leaving the issue of
hyper-parameter tuning for future work.  The diversity of the
synthesized operators stems from the different fitness functions used
in the generation process.  Every result presented herein was achieved
with the following fixed hyper-parameter values:

\begin{itemize}
\item population size equal to \textit{1000} and elite pool of size
  \textit{10};
\item \textit{50} generations;
\item \textit{subtree} cross-over operator with probability
  \textit{0.75};
\item \textit{subtree} mutation operator applied \textit{2} times to
  every individual;
\item \textit{tournament} selection strategy of size \textit{2};
\item \textit{PI-grow}~\cite{fagan_exploring_2016} initialization method;
\item generated syntax tree limited to maximum depth \textit{90}.
\end{itemize} 

Experiments were conducted on a Debian machine, featuring four 16-core
Opteron 6376 CPUs and 128GB ECC RAM.  Effectively, it allowed to
evaluate the population in parallel with 64 separate processes.  On
the software side, a 4.9.130 Linux kernel has been used, together with
Grammar Evolution implemented by the current development version of
the PonyGE2 framework~\cite{fenton_ponyge2:_2017} (commit
\textit{06b1a8c}) running on top of CPython 3.5.3.
Figure~\ref{fig:experiment} (on page~\pageref{fig:experiment})
presents the flow of data in the experiment.

\subsection{Problem Representation}
\label{sub:problem}

All experiments were conducted using the standard Constraint
Programming representation of the TSP problem based on the
\textit{circuit} global constraint, defined over an array of
\textit{next} variables, corresponding to the list of edges making up
the TSP route.  Compared to the more basic \textit{alldifferent}
constraint, this one reveals more information about the problem
structure.

As NDL supports only binary constraints, the global constraint
\textit{circuit} had to be decomposed by adding an additional array of
variables \textit{order}, representing the cities visited in the order
defined by the edges, starting at the first city:

\[   
\textit{order}[i] = 
\begin{cases}
i = 1, & 1 \\
i > 1, & next[order[i-1]]
\end{cases}
\]


It is worth noting that it is unnecessary to handle the \textit{order}
variables explicitly, as they depend directly on the \textit{next}
array.  The Noodle framework updates the values of such variables
during constraint propagation and, at the moment, they are not
accessible from the neighborhood operator itself.

There are three types of constraints involved in the problem
decomposition.  Assuming that $\mathbb{I}$ is a set of city indexes:
\begin{itemize}
\item \textit{all\_diff\_next} --- every pair of the \textit{next}
  variables has to be different:
  $$\forall i,j \in \mathbb{I}: i \neq j \implies next[i] \neq
  next[j]$$
\item \textit{all\_diff\_order} --- every pair of the \textit{order}
  variables has to be different:
  $$\forall i,j \in \mathbb{I}:  i \neq j \implies order[i] \neq order[j]$$
\item \textit{self\_diff\_next} --- a redundant unary constraint
  preventing self-loops by stating that variable of type \textit{next}
  should not point at itself:
  $$\forall i \in \mathbb{I}: \textit{next}[i] \neq i$$
\end{itemize}

While all the constraints take part in the fitness function as
explained in Section~\ref{sub:fitness}, \textit{all\_diff\_next} may
be also used to explore the constraint graph via the \textit{Least
  Fixpoint Operator} combinator introduced in Section~\ref{sec:ndl}.
As \textit{next} variables use the same set of values for both domain
and index set, they make take part in the \textit{Iterator}
combinator, and may be encoded with only two sets of terminals:
$\{$T0, T1, \dots $\}$ representing the variables and $\{$D0, D1,
\dots $\}$ representing both indexes and values.

The evaluation instances used in the experiment consisted of two basic TSP 
problems, containing respectively six and seven cities, and four admissible 
solutions, two per problem. 

\subsection{Synthesized Neighborhood Operators}

\begin{figure}[!ht]
  \centering 
  \subfloat{\includegraphics[width=\linewidth]{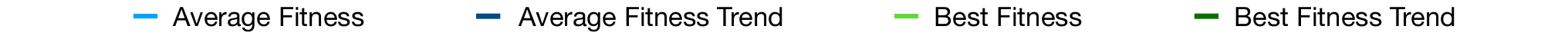}}\\
  \setcounter{subfigure}{0}
  \subfloat[][\textit{2-opt} 
  operator.]{\includegraphics[width=.4\linewidth]{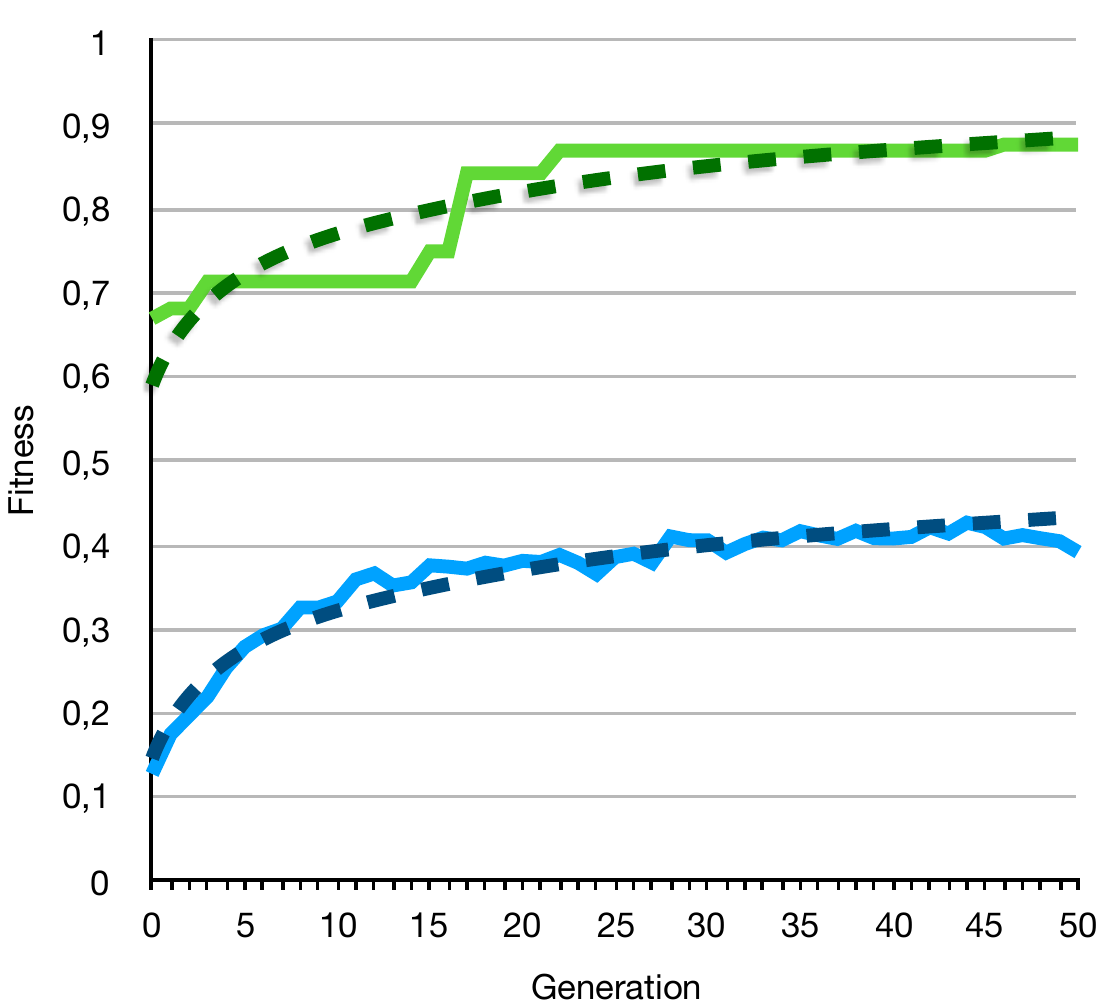}}\quad
  \subfloat[][\textit{Basic 3-opt} 
  operator.]{\includegraphics[width=.4\linewidth]{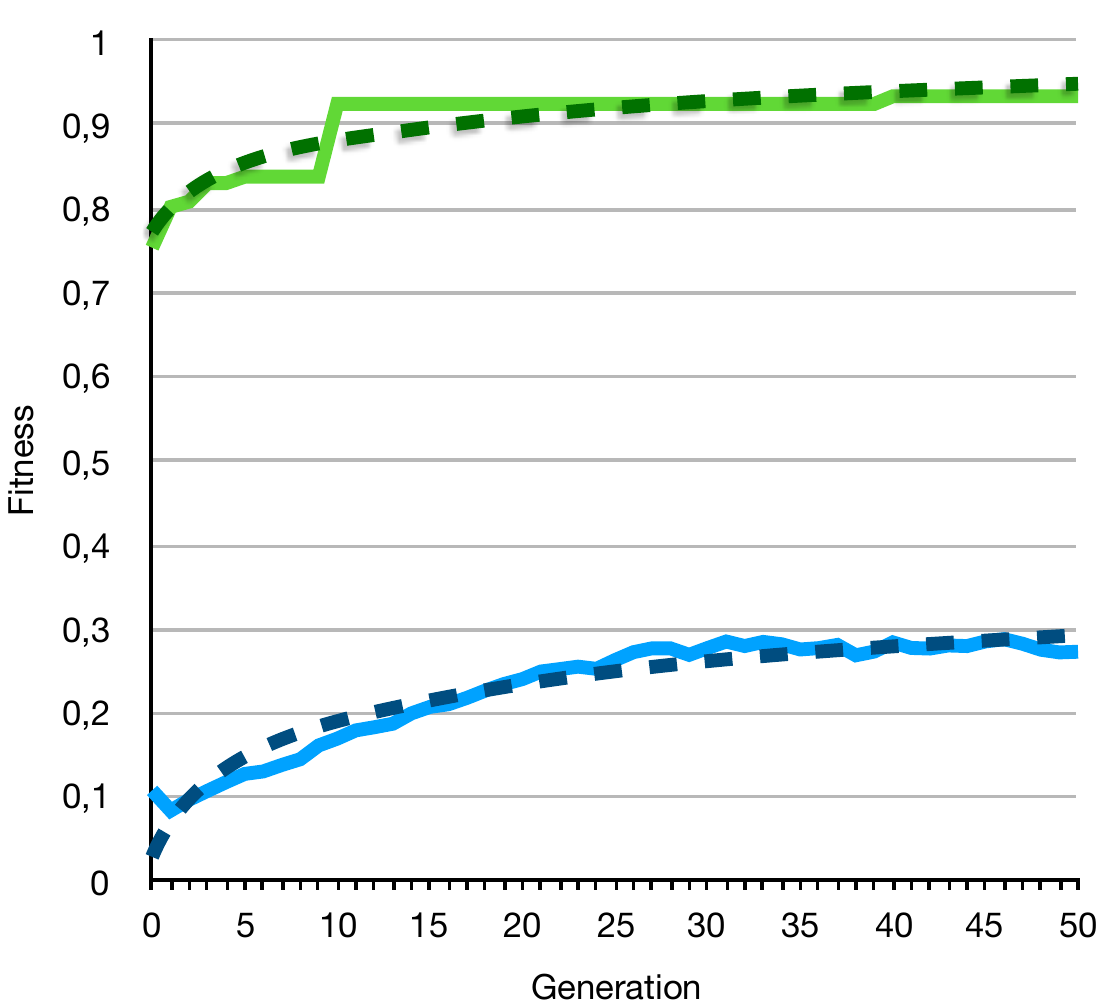}}\\
  \subfloat[][\textit{3-swap} 
  operator.]{\includegraphics[width=.4\linewidth]{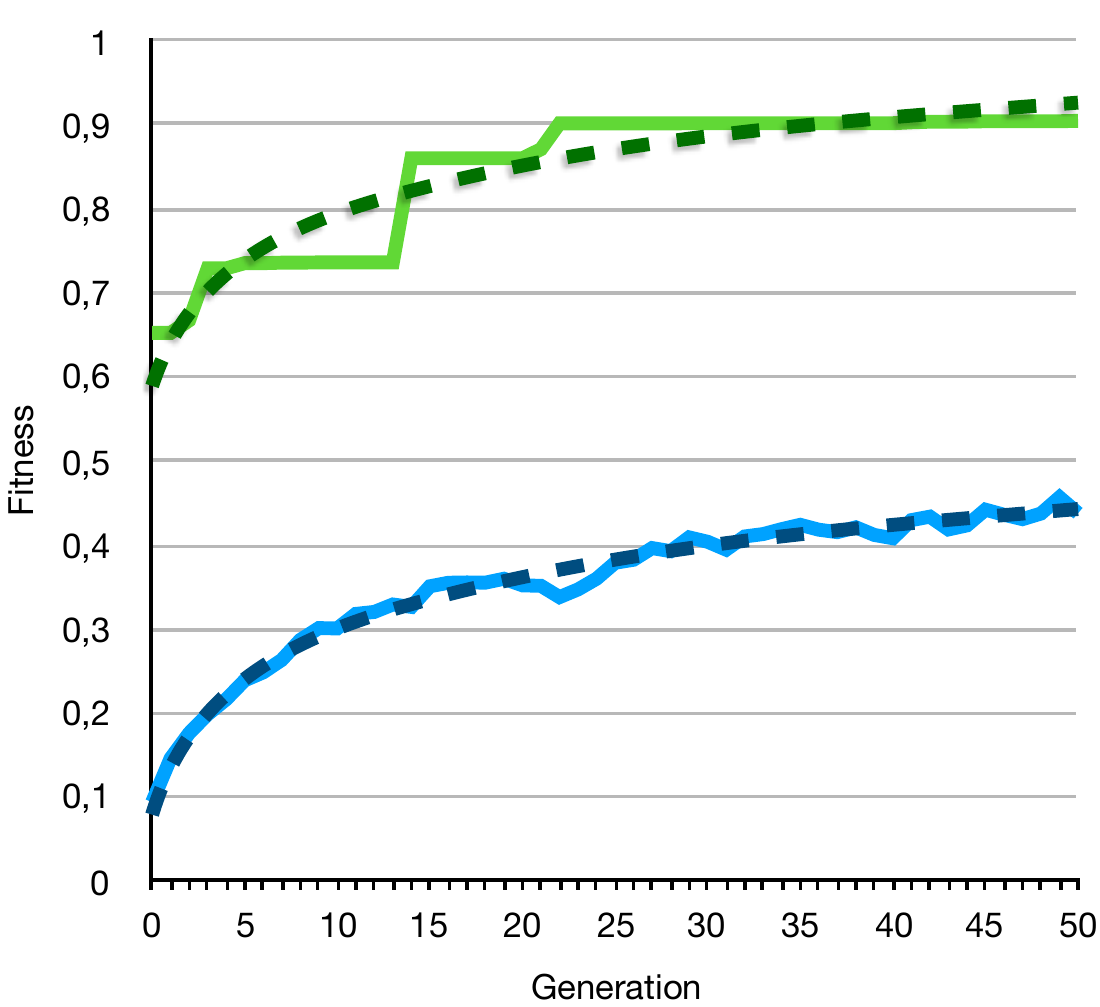}}\quad
  \subfloat[][\textit{Even swap} 
  operator]{\includegraphics[width=.4\linewidth]{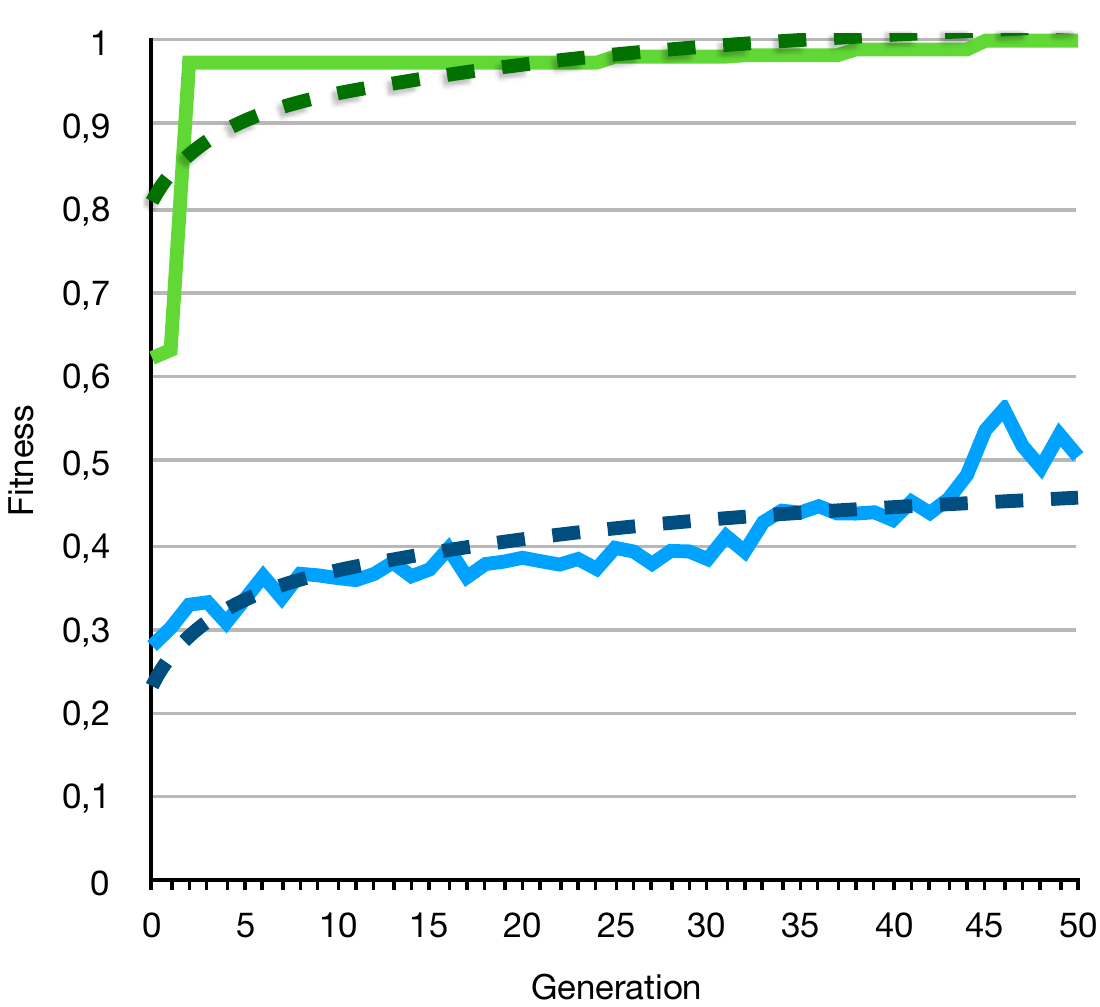}}\\
  \caption{Fitness improvement during synthesis process. All plots were 
  normalized into $(0, 1)$ range.}
  \label{fig:fitness-plot}
\end{figure} 

The experiments we carried out were designed to examine the impact of
various fitness functions on synthesis performance and locally optimal
neighborhood operators.  It turned out we were able to synthesize four
distinct interesting operators --- two of which already known in the
literature as \emph{2-opt} and
\emph{3-opt}~\cite{johnson_traveling_1997} and two others which, to
the best of our knowledge, lack a proper name and were first time baptized in this 
paper. 
Figure~\ref{fig:fitness-plot} shows how fitness of the population
progressed during the evolutionary synthesis process of each operator.
The noticeable gap between best and average fitness is explained by
the destructive influence of the genetic operators and the replacement
strategy being used --- most of the individuals in each generation are
partly random and happen to be semantically incorrect which leads to
very low fitness score for the neighborhood quality.  It is also worth
noting that, while the average fitness tends to continuously improve,
the best individuals change in a more discrete manner and correspond
to novel neighborhood operators.

Because of space constraints, we only detail the 2-opt case, resorting
to a brief description of the other results.

\subsubsection{2-opt Neighborhood}
\label{subsub:2opt}

The 2-opt operator is a widely known and efficient neighborhood
operator used in the traveling salesman (TSP) and related problems.
Given the directed graph representation of the salesman route, there
are two steps involved in a single 2-opt move.  First, two arcs are
removed from the route and replaced with two new arcs, connecting the
corresponding source and sink nodes of the original arcs.  Afterward,
in order to keep the directed graph cyclic, one needs to reverse the
direction of a single path connecting the new arcs.  There are two
important properties to this process:
\begin{itemize}
\item it keeps the configuration admissible, i.e.~if the initial
  configuration was a Hamiltonian cycle, this will remain true for the
  neighbor configuration.
\item the number of reassigned variables varies with the choice of the
  removed edges.
\end{itemize}

\noindent
The fitness function we used in order to synthesize the 2-opt operator
was designed as the following weighted sum:

\begin{equation}
\Phi_\textit{2-opt} =  \phi_\textit{code} + 2 \times \left(
\phi_\textit{NMSS} + 
\phi_\textit{sat} +
\phi_\textit{size} \times
\phi_\textit{unique} 
\times
\phi_\textit{var} \right)
\end{equation}

Where the fitness components $\phi$ were defined in Section
\ref{sub:fitness} and $\Phi_\textit{2-opt} \in \left(0, 7\right)$.
The $\phi_\textit{unique}$ component plays a crucial role as it
penalizes neighborhood operators with a constant number of reassigned
variables, like the basic 3-opt operator described in the
Section~\ref{sub:other}.

Listing~\ref{lst:2opt-raw} contains the 2-opt operator as synthesized
by the Noodle framework.  It is worth noting that lines \texttt{2} and
\texttt{3} do not have any impact on the inferred neighborhood
(variables \texttt{T2}, \texttt{D1}, \texttt{D3} are not connected to
any proper modifier).  As previously mentioned (see
Section~\ref{sub:fitness})) they are treated as introns and therefore
get pruned away before the evaluation.  Line \texttt{4} is also an
intron, but for a different reason --- the \texttt{D2} variable is not
bound at the time and the whole comparison would lead to a run-time
exception.  A simplified version of the same operator is shown in
Listing~\ref{lst:2opt-opt}: it is this code that is tested against the
fitness function.  Still, even when pruned out, introns leave a
negative impact on the $\phi_\textit{code}$ fitness component.

\begin{listing}
\begin{lstlisting}
1.  constraint(all_diff_next, T0, T1) $\land$
2.  variable(D1, T2) $\land$
3.  variable(D3, T2) $\land$
4.  D2 < D3 $\land$
5.  iterate(T0, T3 - T4, 
5.1     constraint(all_diff_next, T4, T1) $\land$
5.2     swap_values(T1, T0) $\land$
5.3     swap_values(T4, T0))
\end{lstlisting}  
 \caption{2-opt Neighborhood operator synthesized by the Noodle framework.}
 \label{lst:2opt-raw}
\end{listing}

\begin{listing}
\begin{lstlisting}
1.  constraint(all_diff_next, T0, T1) $\land$
2.  iterate(T3 - T4, T0, (
2.1     constraint(all_diff_next, T4, T1) $\land$
2.2     swap_values(T1, T0) $\land$
2.3     swap_values(T4, T0)))
\end{lstlisting}
\caption{2-opt operator after removing introns and singleton variables.}
\label{lst:2opt-opt}
\end{listing}

Remarkably, the synthesized operator does not resemble any common
2-opt implementation and requires an explanation on its workings.  As
in the classic 2-opt we can split the process in two parts (line
numbers refer here to Listing~\ref{lst:2opt-opt}):

\begin{enumerate}

\item Line \texttt{1} binds NDL variables \texttt{T0} and \texttt{T1}
  to two problem variables involved in a \texttt{all\_diff\_next}
  constraint.  Effectively, this selects two different variables
  corresponding to two distinct nodes from the TSP route.

\item Line \texttt{2} iterates through the \texttt{next} variables
  starting from \texttt{T0}:
  \begin{itemize}

  \item Line \texttt{2.1} states that iteration will continue only if
    variables \texttt{T4} and \texttt{T1} are connected with the
    \texttt{all\_diff\_next} constraint, i.e.~only when they are
    different variables (it means the same as \texttt{T4
      \textbackslash= T1}).

  \item Line \texttt{2.2} swaps the values of \texttt{T0}, and
    \texttt{T1}, effectively removing two arcs from the route, and
    reassembling it into two sub-routes.

  \item Line \texttt{2.3} creates a new route from the sub-routes
    created in the previous step.

  \item The iterator goes through all the edges in the path connecting
    \texttt{T0} and \texttt{T1}; then terminates due to the line
    \texttt{2.1} being unsatisfied, i.e.  because \texttt{T4 = T1}.

  \end{itemize}

\end{enumerate} 

\begin{figure}
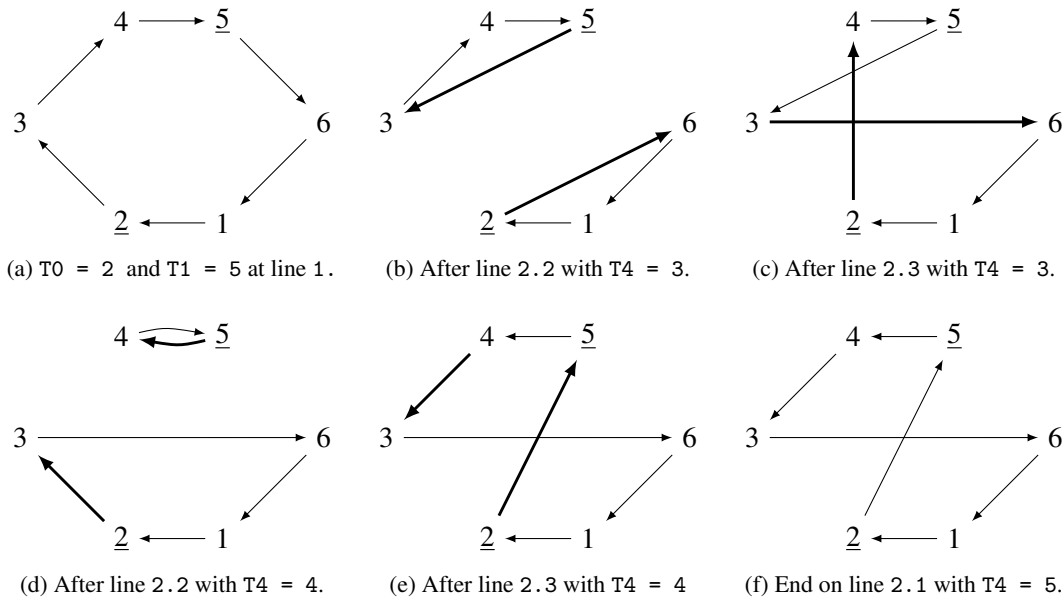

  \centering 
  \subfloat[][\texttt{T0 = 2 }and \texttt{T1 = 5} at line \texttt{1.} 
  ]{\includegraphics[width=.28\linewidth]{2opt_1.tikz}}\quad
  \subfloat[][After line \texttt{2.2} with \texttt{T4 = 
    3}.]{\includegraphics[width=.28\linewidth]{tikz/2opt_2.tikz}}\quad
  \subfloat[][After line \texttt{2.3} with \texttt{T4 = 
    3}.]{\includegraphics[width=.28\linewidth]{tikz/2opt_3.tikz}}\\
  \subfloat[][After line \texttt{2.2} with \texttt{T4 = 
    4}.]{\includegraphics[width=.28\linewidth]{tikz/2opt_4.tikz}}\quad
  \subfloat[][After line \texttt{2.3} with \texttt{T4 = 
    4}]{\includegraphics[width=.28\linewidth]{tikz/2opt_5.tikz}}\quad
  \subfloat[][End on line \texttt{2.1} with \texttt{T4 = 
    5}.]{\includegraphics[width=.28\linewidth]{tikz/2opt_6.tikz}}\\
  \caption{Sequence of consecutive configurations in the 2-opt neighborhood as 
    defined in Listing \ref{lst:2opt-opt}.}
  \label{fig:2opt-states}
\end{figure}

Figure~\ref{fig:2opt-states} presents effects of the synthesized 2-opt
operator, performed on a simple 6-node traveling salesman route.

\subsection{Other Results}
\label{sub:other}
We synthesized three other neighborhood operators: basic 3-opt, even
swap and 3-swap.

\paragraph{Basic 3-opt} is a simplified operator from the 3-opt
family; one that not requires reversing any path in the graph. Given
two different \textit{next} variables \texttt{T0}, \texttt{T1} and
\texttt{T2} being \texttt{T1}'s successor, their values gets swapped
such that to \texttt{T0} is assigned an old value of \texttt{T1}, to
\texttt{T1} value of \texttt{T2} and to \texttt{T2} value of
\texttt{T0}.  This operator is a strong attractor when the fitness
function does not reward operators with a varying amount of changes:

\begin{equation}
\Phi_\textit{3-opt} =  \phi_\textit{code} + 2 \times \left(
\phi_\textit{NMSS} + 
\phi_\textit{sat} +
\phi_\textit{size} \times
\phi_\textit{unique} \right)
\end{equation}

\paragraph{Even swap} is a very simple neighborhood operator that
keeps the solution admissible only when the route length is even.
Starting at a single \textit{next} variable \texttt{T0}, it iterates
backwards over other variables, swapping values of \texttt{T0} and its
predecessors.  Every odd swap breaks the route into two independent
cycles and every even swap reassembles them back into a whole.  Given
the even number of the variables, the process always ends with an
admissible configuration, quite different from the initial one.  This
operator was synthesized as a result of over-fitting to the test data
containing only routes of even length.  The fitness function being
used was the same as in the 3-opt case.

\paragraph{3-swap} is a complex operator composed of two operators:
\textit{basic 3-opt} and an \textit{even swap}, improved with checks
to work on routes of any length.  This operator is a result of
over-fitting due to the amount of features included in the fitness
function:

\begin{align*}
\Phi_\textit{3-swap} = \phi_\textit{code} + 2 \times (&
\phi_\textit{NMSS} + 
\phi_\textit{sat} +
0.6 \times 
\phi_\textit{size} \times
\phi_\textit{unique} 
\times
\phi_\textit{var}+ \\& 
0.1 \times 
\phi_\textit{size} \times
\phi_\textit{unique}
+ 0.1 \times 
\phi_\textit{unique} 
\times
\phi_\textit{var}
+ 0.1 \times 
\phi_\textit{size} 
\times
\phi_\textit{var}+\\&
0.05 \times \phi_\textit{size} + 0.05 \times \phi_\textit{amount})
\end{align*}


\section{Conclusion}

In this paper we have presented a program synthesis framework capable
of finding problem-specific Local Search neighborhoods based on the
Constraint Programming model structure.  To achieve this goal, a novel
neighborhood representation scheme, called Neighborhood Description
Language or NDL, has been introduced.  We have shown that NDL
statements may be compiled to Prolog code and executed by means of
Noodle runtime.  To synthesize the NDL programs, a Grammar Evolution
algorithm has been proposed along with adequate fitness functions and
a method for creating admissible formal grammars.  The system
described herein has been evaluated on the Traveling Salesman Problem
(TSP) and proved able of reproducing efficient neighborhood operators,
commonly used among the optimization community, among others.

We plan to research and experiment so as to cover other discrete
optimization problems, including domains with no known efficient Local
Search strategy.  It will require more detailed insight into the
involved hyper-parameters and the algorithm configuration methods.
All the novel results, including TSP neighborhoods presented in this
paper, need to be analyzed in terms of their runtime solving
efficiency, using public test instances.  Such evaluation may be
interleaved with the synthesis process, guiding it further into
promising search space areas.


\bibliographystyle{eptcs}
\bibliography{bibliography/auto,bibliography/ndl,bibliography/problems,bibliography/misc,bibliography/cp_modelling,bibliography/search_modelling,bibliography/genetic_programming,bibliography/inductive_programming,bibliography/parallel_search}

\begin{thebibliography}{10}
\providecommand{\bibitemdeclare}[2]{}
\providecommand{\surnamestart}{}
\providecommand{\surnameend}{}
\providecommand{\urlprefix}{Available at }
\providecommand{\url}[1]{\texttt{#1}}
\providecommand{\href}[2]{\texttt{#2}}
\providecommand{\urlalt}[2]{\href{#1}{#2}}
\providecommand{\doi}[1]{doi:\urlalt{http://dx.doi.org/#1}{#1}}
\providecommand{\bibinfo}[2]{#2}

\bibitemdeclare{inproceedings}{akgun_framework_2018}
\bibitem{akgun_framework_2018}
\bibinfo{author}{\surnamestart Özgür Akgün\surnameend},
  \bibinfo{author}{Saad \surnamestart Attieh\surnameend},
  \bibinfo{author}{Ian~P. \surnamestart Gent\surnameend},
  \bibinfo{author}{Christopher \surnamestart Jefferson\surnameend},
  \bibinfo{author}{Ian \surnamestart Miguel\surnameend}, \bibinfo{author}{Peter
  \surnamestart Nightingale\surnameend}, \bibinfo{author}{András~Z.
  \surnamestart Salamon\surnameend}, \bibinfo{author}{Patrick \surnamestart
  Spracklen\surnameend} \& \bibinfo{author}{James \surnamestart
  Wetter\surnameend} (\bibinfo{year}{2018}): \emph{\bibinfo{title}{A Framework
  for Constraint Based Local Search using Essence}}.
\newblock In: {\sl \bibinfo{booktitle}{Proceedings of the Twenty-Seventh
  International Joint Conference on Artificial Intelligence, {IJCAI-18}}},
  \bibinfo{publisher}{International Joint Conferences on Artificial
  Intelligence Organization}, pp. \bibinfo{pages}{1242--1248},
  \doi{10.24963/ijcai.2018/173}.

\bibitemdeclare{incollection}{angeline_advances_1994}
\bibitem{angeline_advances_1994}
\bibinfo{author}{Peter~J. \surnamestart Angeline\surnameend}
  (\bibinfo{year}{1994}): \emph{\bibinfo{title}{Genetic programming and
  emergent intelligence}}.
\newblock In \bibinfo{editor}{Kenneth~E. \surnamestart Kinnear\surnameend,
  Jr.}, editor: {\sl \bibinfo{booktitle}{Advances in {Genetic} {Programming}}},
  \bibinfo{publisher}{MIT Press}, \bibinfo{address}{Cambridge, MA, USA}, pp.
  \bibinfo{pages}{75--97}.
\newblock \urlprefix\url{http://dl.acm.org/citation.cfm?id=185984.185992}.

\bibitemdeclare{article}{bjordal_constraint-based_2015}
\bibitem{bjordal_constraint-based_2015}
\bibinfo{author}{Gustav \surnamestart Björdal\surnameend},
  \bibinfo{author}{Jean-Noël \surnamestart Monette\surnameend},
  \bibinfo{author}{Pierre \surnamestart Flener\surnameend} \&
  \bibinfo{author}{Justin \surnamestart Pearson\surnameend}
  (\bibinfo{year}{2015}): \emph{\bibinfo{title}{A constraint-based local search
  backend for {MiniZinc}}}.
\newblock {\sl \bibinfo{journal}{Constraints}}
  \bibinfo{volume}{20}(\bibinfo{number}{3}), pp. \bibinfo{pages}{325--345},
  \doi{10.1007/s10601-015-9184-z}.

\bibitemdeclare{inproceedings}{elsayed_synthesis_2011}
\bibitem{elsayed_synthesis_2011}
\bibinfo{author}{Samir A.~Mohamed \surnamestart Elsayed\surnameend} \&
  \bibinfo{author}{Laurent \surnamestart Michel\surnameend}
  (\bibinfo{year}{2011}): \emph{\bibinfo{title}{Synthesis of {Search}
  {Algorithms} from {High}-{Level} {CP} {Models}}}.
\newblock In \bibinfo{editor}{Jimmy \surnamestart Lee\surnameend}, editor: {\sl
  \bibinfo{booktitle}{Principles and {Practice} of {Constraint} {Programming}
  – {CP} 2011}}, \bibinfo{series}{Lecture {Notes} in {Computer} {Science}},
  \bibinfo{publisher}{Springer Berlin Heidelberg}, pp.
  \bibinfo{pages}{256--270}, \doi{10.1007/978-3-642-23786-7\_21}.

\bibitemdeclare{inbook}{elsken_neural_2019}
\bibitem{elsken_neural_2019}
\bibinfo{author}{Thomas \surnamestart Elsken\surnameend},
  \bibinfo{author}{Jan~Hendrik \surnamestart Metzen\surnameend} \&
  \bibinfo{author}{Frank \surnamestart Hutter\surnameend}
  (\bibinfo{year}{2019}): \emph{\bibinfo{title}{Neural Architecture Search}},
  pp. \bibinfo{pages}{63--77}.
\newblock \bibinfo{publisher}{Springer International Publishing},
  \bibinfo{address}{Cham}, \doi{10.1007/978-3-030-05318-5\_3}.

\bibitemdeclare{inproceedings}{fagan_exploring_2016}
\bibitem{fagan_exploring_2016}
\bibinfo{author}{D.~\surnamestart Fagan\surnameend},
  \bibinfo{author}{M.~\surnamestart Fenton\surnameend} \&
  \bibinfo{author}{M.~\surnamestart O'Neill\surnameend} (\bibinfo{year}{2016}):
  \emph{\bibinfo{title}{Exploring position independent initialisation in
  grammatical evolution}}.
\newblock In: {\sl \bibinfo{booktitle}{2016 {IEEE} {Congress} on {Evolutionary}
  {Computation} ({CEC})}}, pp. \bibinfo{pages}{5060--5067},
  \doi{10.1109/CEC.2016.7748331}.

\bibitemdeclare{article}{fenton_ponyge2:_2017}
\bibitem{fenton_ponyge2:_2017}
\bibinfo{author}{Michael \surnamestart Fenton\surnameend},
  \bibinfo{author}{James \surnamestart McDermott\surnameend},
  \bibinfo{author}{David \surnamestart Fagan\surnameend},
  \bibinfo{author}{Stefan \surnamestart Forstenlechner\surnameend},
  \bibinfo{author}{Michael \surnamestart O'Neill\surnameend} \&
  \bibinfo{author}{Erik \surnamestart Hemberg\surnameend}
  (\bibinfo{year}{2017}): \emph{\bibinfo{title}{{PonyGE}2: {Grammatical}
  {Evolution} in {Python}}}.
\newblock {\sl \bibinfo{journal}{Proceedings of the Genetic and Evolutionary
  Computation Conference Companion on - GECCO '17}}, pp.
  \bibinfo{pages}{1194--1201}, \doi{10.1145/3067695.3082469}.
\newblock \urlprefix\url{http://arxiv.org/abs/1703.08535}.
\newblock \bibinfo{note}{ArXiv: 1703.08535}.

\bibitemdeclare{inproceedings}{he_solution_2012}
\bibitem{he_solution_2012}
\bibinfo{author}{Jun \surnamestart He\surnameend}, \bibinfo{author}{Pierre
  \surnamestart Flener\surnameend} \& \bibinfo{author}{Justin \surnamestart
  Pearson\surnameend} (\bibinfo{year}{2012}): \emph{\bibinfo{title}{Solution
  {Neighbourhoods} for {Constraint}-directed {Local} {Search}}}.
\newblock In: {\sl \bibinfo{booktitle}{Proceedings of the 27th {Annual} {ACM}
  {Symposium} on {Applied} {Computing}}}, \bibinfo{series}{{SAC} '12},
  \bibinfo{publisher}{ACM}, \bibinfo{address}{New York, NY, USA}, pp.
  \bibinfo{pages}{74--79}, \doi{10.1145/2245276.2245294}.

\bibitemdeclare{article}{januario_new_2016}
\bibitem{januario_new_2016}
\bibinfo{author}{Tiago \surnamestart Januario\surnameend} \&
  \bibinfo{author}{Sebastián \surnamestart Urrutia\surnameend}
  (\bibinfo{year}{2016}): \emph{\bibinfo{title}{A new neighborhood structure
  for round robin scheduling problems}}.
\newblock {\sl \bibinfo{journal}{Computers \& Operations Research}}
  \bibinfo{volume}{70}, pp. \bibinfo{pages}{127--139},
  \doi{10.1016/j.cor.2015.12.016}.

\bibitemdeclare{incollection}{johnson_traveling_1997}
\bibitem{johnson_traveling_1997}
\bibinfo{author}{David~S. \surnamestart Johnson\surnameend} \&
  \bibinfo{author}{Lyle~A. \surnamestart McGeoch\surnameend}
  (\bibinfo{year}{1997}): \emph{\bibinfo{title}{The traveling salesman problem:
  {A} case study in local optimization}}.
\newblock In \bibinfo{editor}{E.~H.~L. \surnamestart Aarts\surnameend} \&
  \bibinfo{editor}{J.~K. \surnamestart Lenstra\surnameend}, editors: {\sl
  \bibinfo{booktitle}{Local {Search} in {Combinatorial} {Optimization}}},
  \bibinfo{publisher}{John Wiley and Sons}, \bibinfo{address}{Chichester,
  United Kingdom}, pp. \bibinfo{pages}{215--310}.

\bibitemdeclare{inproceedings}{kaul_autolearn_2017}
\bibitem{kaul_autolearn_2017}
\bibinfo{author}{A.~\surnamestart Kaul\surnameend},
  \bibinfo{author}{S.~\surnamestart Maheshwary\surnameend} \&
  \bibinfo{author}{V.~\surnamestart Pudi\surnameend} (\bibinfo{year}{2017}):
  \emph{\bibinfo{title}{{AutoLearn} — {Automated} {Feature} {Generation} and
  {Selection}}}.
\newblock In: {\sl \bibinfo{booktitle}{2017 {IEEE} {International} {Conference}
  on {Data} {Mining} ({ICDM})}}, pp. \bibinfo{pages}{217--226},
  \doi{10.1109/ICDM.2017.31}.

\bibitemdeclare{misc}{marte_yuck_2017}
\bibitem{marte_yuck_2017}
\bibinfo{author}{Michael \surnamestart Marte\surnameend}
  (\bibinfo{year}{2017}): \emph{\bibinfo{title}{Yuck is a constraint-based
  local-search solver with {FlatZinc} interface}}.
\newblock \urlprefix\url{https://github.com/informarte/yuck}.
\newblock \bibinfo{note}{Accessed 2019/05/13}.

\bibitemdeclare{inproceedings}{meijer_functional_1991}
\bibitem{meijer_functional_1991}
\bibinfo{author}{Erik \surnamestart Meijer\surnameend},
  \bibinfo{author}{Maarten \surnamestart Fokkinga\surnameend} \&
  \bibinfo{author}{Ross \surnamestart Paterson\surnameend}
  (\bibinfo{year}{1991}): \emph{\bibinfo{title}{Functional programming with
  bananas, lenses, envelopes and barbed wire}}.
\newblock In \bibinfo{editor}{John \surnamestart Hughes\surnameend}, editor:
  {\sl \bibinfo{booktitle}{Functional Programming Languages and Computer
  Architecture}}, \bibinfo{publisher}{Springer Berlin Heidelberg},
  \bibinfo{address}{Berlin, Heidelberg}, pp. \bibinfo{pages}{124--144},
  \doi{10.1007/3540543961\_7}.

\bibitemdeclare{incollection}{oneil_grammatical_2003}
\bibitem{oneil_grammatical_2003}
\bibinfo{author}{Michael \surnamestart O’Neil\surnameend} \&
  \bibinfo{author}{Conor \surnamestart Ryan\surnameend} (\bibinfo{year}{2003}):
  \emph{\bibinfo{title}{Grammatical {Evolution}}}.
\newblock In: {\sl \bibinfo{booktitle}{Grammatical {Evolution}}},
  \bibinfo{series}{Genetic {Programming} {Series}},
  \bibinfo{publisher}{Springer, Boston, MA}, pp. \bibinfo{pages}{33--47},
  \doi{10.1007/978-1-4615-0447-4\_4}.

\bibitemdeclare{article}{shahriari_taking_2016}
\bibitem{shahriari_taking_2016}
\bibinfo{author}{B.~\surnamestart Shahriari\surnameend},
  \bibinfo{author}{K.~\surnamestart Swersky\surnameend},
  \bibinfo{author}{Z.~\surnamestart Wang\surnameend}, \bibinfo{author}{R.~P.
  \surnamestart Adams\surnameend} \& \bibinfo{author}{N.~de \surnamestart
  Freitas\surnameend} (\bibinfo{year}{2016}): \emph{\bibinfo{title}{Taking the
  {Human} {Out} of the {Loop}: {A} {Review} of {Bayesian} {Optimization}}}.
\newblock {\sl \bibinfo{journal}{Proceedings of the IEEE}}
  \bibinfo{volume}{104}(\bibinfo{number}{1}), pp. \bibinfo{pages}{148--175},
  \doi{10.1109/JPROC.2015.2494218}.

\bibitemdeclare{inproceedings}{slazynski_towards_2019}
\bibitem{slazynski_towards_2019}
\bibinfo{author}{Mateusz \surnamestart \'{S}la\.{z}y\'{n}ski\surnameend},
  \bibinfo{author}{Salvador \surnamestart Abreu\surnameend} \&
  \bibinfo{author}{Grzegorz~J. \surnamestart Nalepa\surnameend}
  (\bibinfo{year}{2019}): \emph{\bibinfo{title}{Towards a Formal Specification
  of Local Search Neighborhoods from a Constraint Satisfaction Problem
  Structure}}.
\newblock In: {\sl \bibinfo{booktitle}{Proceedings of the Genetic and
  Evolutionary Computation Conference Companion}}, \bibinfo{series}{GECCO '19},
  \bibinfo{publisher}{ACM}, \bibinfo{address}{New York, NY, USA}, pp.
  \bibinfo{pages}{137--138}, \doi{10.1145/3319619.3321968}.

\bibitemdeclare{book}{van_hentenryck_constraint-based_2005}
\bibitem{van_hentenryck_constraint-based_2005}
\bibinfo{author}{Pascal \surnamestart Van~Hentenryck\surnameend} \&
  \bibinfo{author}{Laurent \surnamestart Michel\surnameend}
  (\bibinfo{year}{2005}): \emph{\bibinfo{title}{Constraint-{Based} {Local}
  {Search}}}.
\newblock \bibinfo{publisher}{The MIT Press}.

\bibitemdeclare{inproceedings}{van_hentenryck_synthesis_2007}
\bibitem{van_hentenryck_synthesis_2007}
\bibinfo{author}{Pascal \surnamestart Van~Hentenryck\surnameend} \&
  \bibinfo{author}{Laurent \surnamestart Michel\surnameend}
  (\bibinfo{year}{2007}): \emph{\bibinfo{title}{Synthesis of {Constraint}-based
  {Local} {Search} {Algorithms} from {High}-level {Models}}}.
\newblock In: {\sl \bibinfo{booktitle}{Proceedings of the 22Nd {National}
  {Conference} on {Artificial} {Intelligence} - {Volume} 1}},
  \bibinfo{series}{{AAAI}'07}, \bibinfo{publisher}{AAAI Press},
  \bibinfo{address}{Vancouver, British Columbia, Canada}, pp.
  \bibinfo{pages}{273--278}.

\bibitemdeclare{inproceedings}{van_hentenryck_traveling_2006}
\bibitem{van_hentenryck_traveling_2006}
\bibinfo{author}{Pascal \surnamestart Van~Hentenryck\surnameend} \&
  \bibinfo{author}{Yannis \surnamestart Vergados\surnameend}
  (\bibinfo{year}{2006}): \emph{\bibinfo{title}{Traveling {Tournament}
  {Scheduling}: {A} {Systematic} {Evaluation} of {Simulated} {Annealling}}}.
\newblock In \bibinfo{editor}{J.~Christopher \surnamestart Beck\surnameend} \&
  \bibinfo{editor}{Barbara~M. \surnamestart Smith\surnameend}, editors: {\sl
  \bibinfo{booktitle}{Integration of {AI} and {OR} {Techniques} in {Constraint}
  {Programming} for {Combinatorial} {Optimization} {Problems}}},
  \bibinfo{series}{Lecture {Notes} in {Computer} {Science}},
  \bibinfo{publisher}{Springer Berlin Heidelberg}, pp.
  \bibinfo{pages}{228--243}, \doi{10.1007/11757375\_19}.

\bibitemdeclare{article}{wielemaker_swi-prolog_2012}
\bibitem{wielemaker_swi-prolog_2012}
\bibinfo{author}{Jan \surnamestart Wielemaker\surnameend}, \bibinfo{author}{Tom
  \surnamestart Schrijvers\surnameend}, \bibinfo{author}{Markus \surnamestart
  Triska\surnameend} \& \bibinfo{author}{Torbjörn \surnamestart
  Lager\surnameend} (\bibinfo{year}{2012}):
  \emph{\bibinfo{title}{{SWI}-{Prolog}}}.
\newblock {\sl \bibinfo{journal}{Theory and Practice of Logic Programming}}
  \bibinfo{volume}{12}(\bibinfo{number}{1-2}), pp. \bibinfo{pages}{67--96},
  \doi{10.1017/S1471068407003237}.

\end{thebibliography}

\label{lastpage}
\end{document}